\documentclass{article}

\usepackage{PRIMEarxiv}

\usepackage[utf8]{inputenc} 
\usepackage[T1]{fontenc}    
\usepackage{hyperref}       
\usepackage{url}            
\usepackage{booktabs}       
\usepackage{amsfonts}       
\usepackage{nicefrac}       
\usepackage{microtype}      
\usepackage{lipsum}
\usepackage{fancyhdr}       
\usepackage{graphicx}       

\usepackage{amsmath,amssymb,amsfonts}
\usepackage{textcomp}
\usepackage{xcolor}
\usepackage{tipa}
\usepackage{rotating}
\usepackage{subcaption}
\usepackage{soul}
\usepackage{multirow}
\usepackage{tablefootnote}
\usepackage{xurl}
\usepackage{float}
\usepackage{pgfplots}

\usepackage{lineno}

\pgfplotsset{width=10cm,compat=1.9}
\usepackage{amssymb}

\graphicspath{{media/}}     

\title{Enhancing Pollinator Conservation towards Agriculture 4.0: Monitoring of Bees through Object Recognition
}

\author{
  Ajay John Alex \\
  Department of Computer Science \\
  Nottingham Trent University \\
  Nottingham\\
   \And
  Chloe M. Barnes \\
  Department of Applied AI and Robotics\\ ACAIRA \\
  Aston University \\
  Birmingham\\
  \And 
  Pedro Machado \\
  Department of Computer Science \\
  Nottingham Trent University \\
  Nottingham\\
    \And 
  Isibor Ihianle \\
  Department of Computer Science \\
  Nottingham Trent University \\
  Nottingham\\
    \And 
  Gábor Markó \\
  Department of Plant Pathology \\
  Hungarian University of Agriculture \\ and Life Sciences \\
  Budapest\\
    \And 
  Martin Bencsik \\
  Department of Physics \\
  Nottingham Trent University \\
  Nottingham\\
    \And 
  Jordan J. Bird \\
  Department of Computer Science \\
  Nottingham Trent University \\
  Nottingham\\
}

\begin{document}
\maketitle

\begin{abstract}
In an era of rapid climate change and its adverse effects on food production, technological intervention to monitor pollinator conservation is of paramount importance for environmental monitoring and conservation for global food security. The survival of the human species depends on the conservation of pollinators. This article explores the use of Computer Vision and Object Recognition to autonomously track and report bee behaviour from images. A novel dataset of 9664 images containing bees is extracted from video streams and annotated with bounding boxes. With training, validation and testing sets (6722, 1915, and 997 images, respectively), the results of the COCO-based YOLO model fine-tuning approaches show that YOLOv5m is the most effective approach in terms of recognition accuracy. However, YOLOv5s was shown to be the most optimal for real-time bee detection with an average processing and inference time of 5.1ms per video frame at the cost of slightly lower ability. The trained model is then packaged within an explainable AI interface, which converts detection events into timestamped reports and charts, with the aim of facilitating use by non-technical users such as expert stakeholders from the apiculture industry towards informing responsible consumption and production.  
\end{abstract}


\section{Introduction}
\label{sec:introduction}
The decline of pollinators, particularly bees, has emerged as a critical concern with adverse effects on global food security. A loss of 1-10\% of biodiversity per decade has also been observed in recent times\cite{kluser2007global}. Various species of bee play a key role in agricultural production, supporting a wide array of crops, including fruits, vegetables, oilseeds and legumes, just to name a few; Animal pollination supports 30\% of global food production\cite{khalifa2021overview}. In the United Kingdom alone, 34\% of all pollination is provided by one species, the European honey bee (\textit{Apis mellifera})\cite{breeze2011pollination}. Factors such as habitat loss and rapid climate change have led to a worrying decline in bee populations around the world, which poses a direct threat to agricultural sustainability. 

During these troubling times, the potential of technological intervention through computational intelligence presents a promising approach to mitigate these problems. The notion of Agriculture 4.0 is largely based on data-centric automation\cite{ampatzidis2020agroview,liu2023survey,fountas2024agriculture,abbasi2022digitization}, which will lead to improvements in agricultural practices in terms of speed and efficiency through the use of technologies such as the Internet of Things (IoT) and edge-based processing, Big Data, Artificial Intelligence (AI) and Machine Learning, as well as Robotics, among others. Precision agriculture applied to apicultural practices can help us autonomously monitor bee behaviour and, in the future, could provide a noninvasive real-time approach to monitor colonies in the long term. This article introduces a novel application of vision-based computational intelligence algorithms towards bee identification and tracking from video streams, which aim to streamline otherwise labour-intensive manual processes. We show that state-of-the-art recognition algorithms such as those within the YOLO suite of research can provide real-time and highly precise bee tracking and that their findings can be distilled into an accessible format for non-technical stakeholders from industry. As shown in \cite{bhuse2022effect} and \cite{isa2022optimizing}, data augmentation and hyperparameter optimisation are key considerations when aiming for approaches that are useful in the real world beyond the laboratory; the novel combination and application of these state-of-the-art approaches, along with the release of our dataset for future interdisciplinary research, aims to facilitate academic research on the conservation of pollinator populations. 

These aforementioned abilities of object recognition algorithms provide timely benefits to saving time and effort in an autonomous monitoring process and further enabling larger-scale bee behaviour analysis. As argued by Ngo et al.\cite{ngo2019real}, real-time imaging plays a critical role in monitoring honeybee behaviours to assess colony health\cite{biesmeijer2006parallel}. This is especially important during a time of great habitat loss, insecticide use, and rapid climate change. In \cite{ngo2019real}. 

In addition to collecting and annotating a novel large-scale dataset, this study also features scientific novelty in the evaluation of several data augmentation and object detection approaches for bee detection. In addition to the data, all algorithms and models are also released as open source. The literature review reveals that many object detection experiments taking place within precision agriculture and biodiversity studies focus primarily on model ability, while this work also explores the potential for real-time execution. The results show that YOLOv5m emerges as the most accurate approach overall, while YOLOv5s exhibits the most promising performance for real-time use in industry. This study also highlights preliminary results on the significant reduction in the inference time for bee detection when using keyframe selection. 

Beyond technical achievements, this work encapsulates a broader vision for the future of sustainable beekeeping and the conservation of bee colonies. In addition to the trained machine learning models released as open source, an explainable AI interface is also produced that infers the predictions made by the model and distils them into meaningful and useful information for key stakeholders. Therefore, the addition of this module aims to bridge the gap between complex computational intelligence algorithms and practical apiculture in the real world. Thus, this approach not only improves the state of the art in beekeeping practice, but also contributes to the global agendas of responsible production, climate action, and the preservation of terrestrial ecosystems. The survival of the human species depends on these natural ecosystems. 

The scientific contributions arising from this comprehensive exploration of object recognition for bee tracking surround both data and vision models. We collect, annotate, and release a dataset of 9,664 images which is available open-source to the research community. In addition, our proposed approach yields several sets of promising results, including the YOLOv5m model, which achieves an mAP@0.5 of 85.6\% at 8.1ms per frame. We also release an explainable AI interface prototype to make the model's findings accessible to non-technical key stakeholders in order to facilitate real-world application and contribute to broader goals of sustainable agriculture and environmental conservation. 

The remainder of this article is as follows. A review of the literature is presented in Section \ref{sec:litreview}, covering the relevant work in apiculture and providing a technical background on the techniques applied in this work. The methodology followed in this work is described in Section \ref{sec:method}, before the results are discussed in Section \ref{sec:results}. Section \ref{sec:conclusion} then suggests future work arising from the findings of this study and finally concludes this article.

\section{Background and Related Work}
This section explores relevant state-of-the-art literature in the field. First, we explore work from an agricultural background before describing the state-of-the-art in object detection techniques, which are relevant to the experiments carried out in this article. 

\label{sec:litreview}
\subsection{Agriculture, Conservation, And Biodiversity}
Bees are an important pollinator in the global ecosystem, helping plant reproduction, and crop quantity and quality. Pollinator services are critical to food production and security; however, bee populations have recently declined. This decline presents a substantial danger to agricultural productivity and the long-term viability of food production\cite{aizen2008long}, which is not only a major contribution to GDP but sustains the survival of the human species\cite{klein2007importance}. 

This decline in pollinators, particularly bees, has become a critical concern with adverse effects on global food security. As described in the introduction, the bee plays a key role in the agricultural production of both food and industrial crops. Animal pollination supports 30\% global food production\cite{khalifa2021overview}, and, beyond food, also plays a significant role in industrial crops. This includes fibres, biofuels, medicinal, material and ornamental plants\cite{komlatskiy2023pollination}. Contributing to the wider discussion of environmental sustainability, pollinators, including bees, play a significant role in the cultivation of biofuel crops\cite{gardiner2010implications,romero2013pollinators}. That is, by facilitating the pollination of biofuel crops such as canola and sunflowers, bees directly improve seed production and thus the availability of these resources, which are critical in sustainable solutions to energy in the global initiative to move away from fossil fuels.

Key stakeholders and beneficiaries, such as farmers in the apicultural field, can implement pollinator-friendly practices that improve agricultural sustainability. For example, the avoidance of harmful insecticides\cite{kremen2007pollination,sanchez2016bee}. Some insecticides can be harmful to bees; symptoms include disruption of navigation, feeding behaviour, and reproduction and immune systems\cite{johansen1983pesticides,pashte2018toxicity}. These negative effects result in a decrease in colony survival rates and contribute to population decline\cite{sanchez2016bee}. Furthermore, climate change has become another risk to bee populations globally\cite{potts2016assessment}; availability of flowers, nesting locations, and general natural dynamics are negatively affected by the observed rise in temperatures and altered precipitation patterns. Autonomous monitoring of bees can provide additional information on their response to these environmental changes, with the benefit of not requiring human presence throughout the whole data collection process. 

This technique contributes to the wider field of environmental monitoring, which is the systematic analysis of the natural environment\cite{kumar2012environmental}. This can enable researchers to understand the current state of the environment, forecast future trends over time, and ultimately assess the impact of human activity on natural systems. Although this article focuses on pollinator behaviour with a focus on the bee, environmental monitoring also encapsulates, but is not limited to, air, water, and soil quality, as well as biodiversity and climate analysis. In the context of this article, monitoring refers to in situ analysis of bee habitats and behavioural patterns to better understand their ecological importance and contribute to conservation efforts with autonomous analysis and data collection. 

Technological intervention to monitor the decline in biodiversity has been promoted in recent times, especially due to the autonomous nature of artificial intelligence and its ability not only to save time and effort, but also to perform tasks for which key stakeholders and beneficiaries simply do not have time to do manually\cite{ratnayake2021towards,stark2023yolo}. To this end, researchers suggest that tracking pollinator behaviour can inform agricultural practices and potentially protect or even improve current crop yield. This could be, for example, information to help optimise crop placement, ensure pollination coverage, and improve the overall efficiency of agricultural systems. 

Bee behaviour is often monitored through various modalities of sensor-based data collection and machine learning, such as in \cite{ramsey2020prediction}, where swarming behaviour was recognised from vibro-acoustic accelerometer data at an accuracy greater than 90\%. Related work in the field includes \cite{magnier2018bee}, which argues that visual alterations, such as background clipping and approximations, can improve pollinator detection through data preprocessing. In \cite{ngo2019real}, researchers suggested the use of techniques that include background subtraction, Kalman filtering, and the Hungarian algorithm to produce a system that can monitor the activity that occurs at the entrance of the beehives. The approach achieved around 93.9\% accuracy ($\pm$1.1\%) compared to manual counting. Similar augmentation techniques were proposed in \cite{ratnayake2021tracking}, where bees in complex outdoor environments were tracked with 86.6\% accuracy using the YOLOv2 object detection model. 

Environmental monitoring has benefitted from data augmentation as a method to improve the ability to perform automated recognition. In \cite{bittner2022generating}, the authors propose that the use of GAN-based synthetic data can help improve the detection of various objects present in forest environments toward forming a more technologically enhanced environmental monitoring system. Kaur et al. proposed a data augmentation strategy to better recognise honeybee disease\cite{kaur2022cnn}, which used a GAN-based approach to improve classification accuracy over conventional augmentation methods. In \cite{de2022image}, researchers showed that traditional image augmentation techniques could improve computer vision-based classification of honeybee subspecies, and Buschbacher et al.\cite{buschbacher2020image} also followed a similar methodology in autonomous recognition of bees. 

The concept of autonomous object detection is at the core of several of the reviewed related works, as well as the main technology behind the approach proposed by this study. This technique is discussed in the following section.

\subsection{Object Detection}
Several related works to this study make use of Bounding Box Object Detection (BBOD) for pollinator monitoring. BBOD is a computer vision technique that predicts the location of a bounding box, or several bounding boxes, around objects of interest. For example, vision systems within an autonomous vehicle aim to draw boundaries or completely segment entities from an image to understand the scene\cite{amit2021object}, such as the detection of an emergency situation where a pedestrian has been detected on a high-speed road.

In \cite{redmon2016you}, a model architecture known as You Only Look Once (YOLO) was proposed, with the aim of implementing real-time BBOD. YOLO is a common deep learning architecture for BBOD that performs the detection task as a single regression problem, directly predicting bounding boxes and class probabilities from full input images in one evaluation. This is unlike many other techniques, where proposals are first generated and then later classified iteratively. YOLO operates a grid-based approach, where predictions are made for each grid cell for bounding box coordinates and two confidence scores (one for the box containing the object and one for the class label of the object itself). More details on the differences between model architectures can be found in \cite{jiang2022review}.

Several metrics are important for object detection, which are featured in this work due to their importance in the state-of-the-art. They include Intersection over Union (IoU), which is a metric used to evaluate the accuracy of an object detection algorithm on a particular dataset. IoU is the overlap between the predicted bounding box and the ground-truth derived from the expert annotation:
\begin{equation}
    \text{IoU} = \frac{\text{Area of Overlap}}{\text{Area of Union}}
\end{equation}
Several different IoU variants, such as GIoU (Generalised IoU) and DIoU (Distance IoU) metrics, have been proposed in the relevant literature to improve the accuracy of object detection algorithms\cite{zheng2020distance}. These metrics consider the size and shape of the predicted and ground-truth bounding boxes, as well as their location and orientation.

Similarly to other machine learning problems, precision and recall are also used as part of the evaluation process. Firstly, precision:
\begin{equation}
    \text{Precision} = \frac{\text{True Positives}}{\text{True Positives} + \text{False Positives}},
\end{equation}
which is the accuracy of the positive predictions. In this case, precision is the proportion of correctly detected bees out of all instances detected as bees that were incorrect. A \textit{True Positive} denotes a predicted bounding box that contains a bee, and a \textit{False Positive} denotes a predicted bounding box that does not contain a bee. 

Recall is also considered:
\begin{equation}
    \text{Recall} = \frac{\text{True Positives}}{\text{True Positives} + \text{False Negatives}}.
\end{equation}
Recall is a measure of the ability of the model to detect all relevant instances. That is, the proportion of true bee instances that were detected by the model out of all instances present in the dataset. \textit{False Negative} in this sense thus denotes a bee in an image that did not receive a predicted bounding box. 

Given the trade-off between precision and recall, the mean Average Precision (mAP) is a useful metric in object detection. mAP@0.5 is the mAP calculated at an IoU threshold of 0.5:
\begin{equation}
    \label{eq:map0.5}
    \text{mAP@0.5} = \frac{1}{N} \sum_{i=1}^{N} \text{AP}_i |_{\text{IoU}=0.5},
\end{equation}
with regards to this work, this means that a prediction is considered a \textit{True Positive} if $\text{IoU} \geq 0.5$.

Beyond a single threshold, the mAP@0.5:0.95 metric measures mAP over multiple thresholds. In this case, $\{0.5, 0.55, 0.6, ..., 0.95\}$: 
\begin{equation}
    \text{mAP@0.5:0.95} = \frac{1}{10} \sum_{t=0.5}^{0.95} \text{mAP}@t.
\end{equation}
The use of multiple thresholds allows for a more rigorous approach, by iterating by increasing levels of strictness for overlap, or $\text{IoU}$.

In addition to the prominent metrics for validation, Region of Interest (RoI) is also an important consideration made by the model. A RoI in the example of bee detection may be a flower patch or even anthers, or the entrance to a hive. Given automatic recognition of RoIs, further analysis can be performed within these areas, and computational resources can be saved by focusing on the parts of the image most likely to contain a bee. Several prominent works in the related literature show that effective RoI detection can significantly reduce the likelihood of false positives from the background\cite{xiang2019end,cores2020roi}. For example, without RoIs, an object detection algorithm trained on a dataset prominent in images of bees collecting pollen may mistakenly classify all flowers as containing bees.

\section{Method}
This section describes the methods used in this work. This includes data collection and preprocessing (with a link to download the data attributed to this work), augmentation machine learning approaches, and finally a description of the method of farmer-facing inference for accessibility. 

\label{sec:method}
\subsection{Data Collection and Preprocessing}

\begin{table}[]
\centering
\caption{Information derived from the dataset.}
\label{tab:dataset-info}
\footnotesize
\begin{tabular}{@{}lr@{}}
\toprule
\textbf{Metric}                          & \textbf{Value} \\ \midrule
Total Number of Images                   & 9,664          \\
Total Number of Bees                     & 13,402         \\
Average Number of Bees per Image         & 1.39           \\
Standard Deviation of Bees per Image     & 1.39           \\
Maximum Number of Bees in a Single Image & 11             \\
Minimum Number of Bees in a Single Image & 0              \\
Number of Images with No Bees            & 1,436          \\ \bottomrule
\end{tabular}
\end{table}

This research project collected an original dataset of digital images named  ``Bee Detection in the Wild" for the purposes of this study. The dataset was collected from various locations in the United Kingdom and annotated by several of the authors of this study, with 9664 annotated images that feature a total of 13402 bounding boxes. On average, there are 1.4 bees per image. Although most of the images (6272) contain one bee, there are a maximum of 11 bees per image. Information derived from the dataset can be found in Table \ref{tab:dataset-info}. Out of the total of 9664 images, 8228 contained at least one instance of a bee. A considerable majority of the images contained one bee, so both the average and standard deviation of bees per image were around 1.39. 

\begin{figure}
    \centering
    \includegraphics[width=\textwidth]{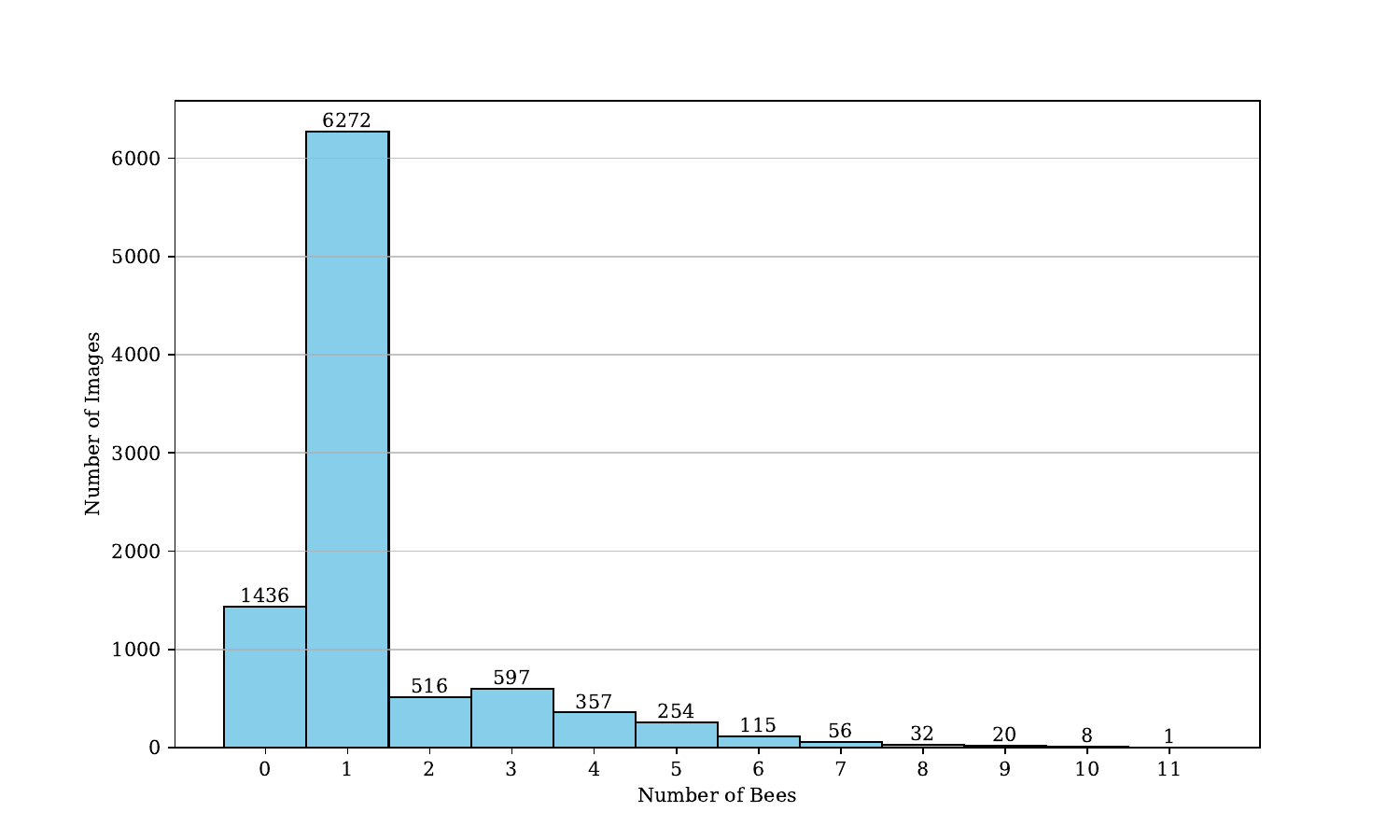}
    \caption{Presence of the number of bees per image within the dataset.}
    \label{fig:bee-counts}
\end{figure}
This can be further observed within Figure \ref{fig:bee-counts}, which shows the distribution of the number of bees per image within the dataset. Higher counts of bees were less frequent, with only one image containing 11 bees, 8 containing 10 bees, and 20 containing 9 bees.

\begin{figure}[ht]
    \centering
    \begin{subfigure}[b]{0.48\textwidth}
        \includegraphics[width=\textwidth]{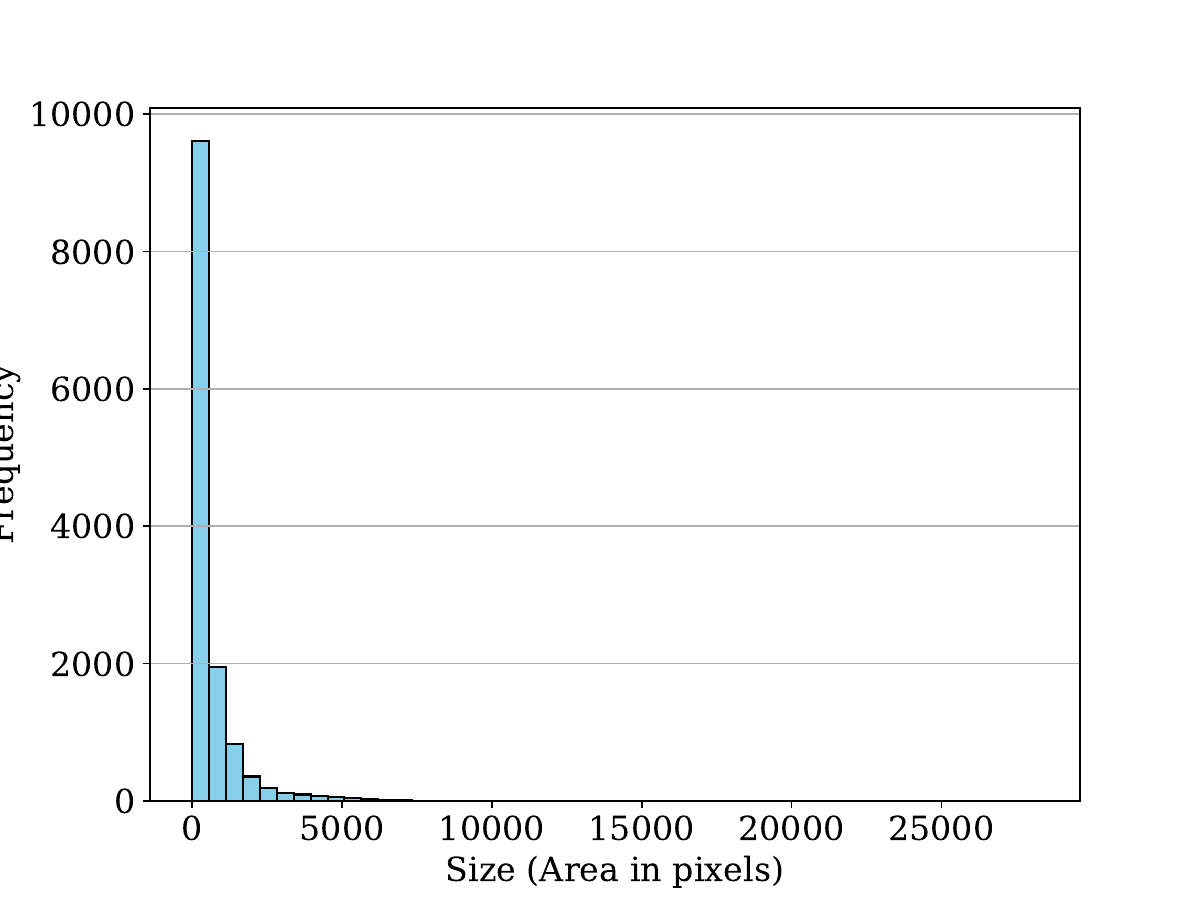}
        \caption{Distribution of Annotation Sizes}
        \label{fig:annotation_sizes}
    \end{subfigure}
    \hfill 
    \begin{subfigure}[b]{0.48\textwidth}
        \includegraphics[width=\textwidth]{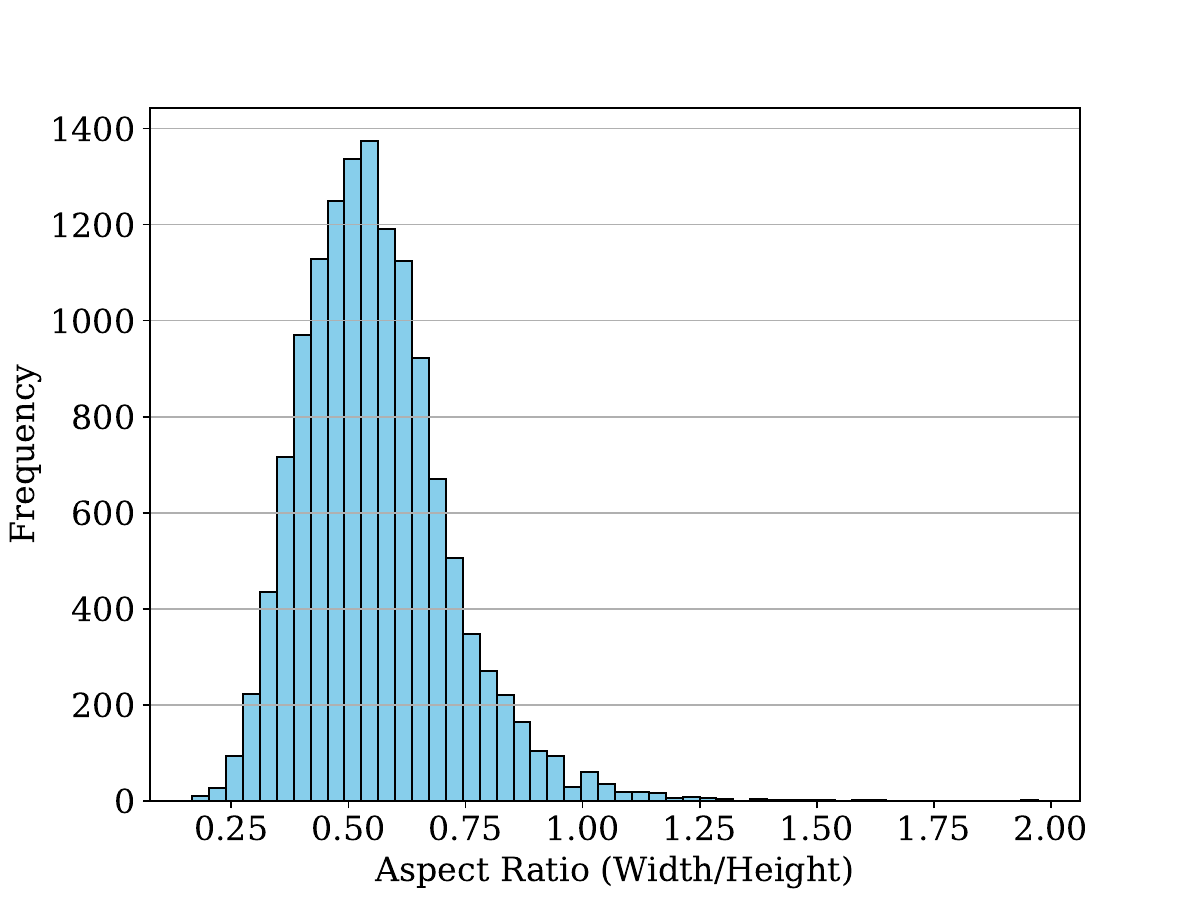}
        \caption{Distribution of Aspect Ratios}
        \label{fig:aspect_ratios}
    \end{subfigure}
    \caption{Annotation Size and Bounding Box Aspect Ratio Distributions within the Dataset.}
    \label{fig:annots-sizes-ratios}
\end{figure}

Figure \ref{fig:annots-sizes-ratios} shows the distribution of annotation sizes and aspect ratios for the entire dataset. Annotation sizes are relatively consistent for most of the dataset, but it should be noted that the dataset does feature a number of images that contain larger bounding boxes. This suggests that while the size of bees is relatively uniform, there are instances of closer views of the bee. The distribution suggests that there is variability in the appearances of bees within the dataset, which may lead to an improved robustness of the object detection algorithm. In terms of bounding box aspect ratios, there exists a much higher diversity within the dataset. This spread suggests that bees are captured in various poses and angles, which, again, may lead to improved robustness of the object detection algorithm, since bees should be detected irrespective of orientation.

During the annotation stage, annotators checked each image individually for those that were deemed to be low quality (such as excessive motion blur), removing them from the dataset. Original images are on average 0.92mp (0.18mp to 2.59mp). The original median image ratio is 1280x720 px. The images in this study are resized to 416px squares for YOLO compatibility, but the original full-resolution images and bounding-box annotations are also released to the research community. 
\begin{figure}[]
    \centering
    \includegraphics[scale=0.21]{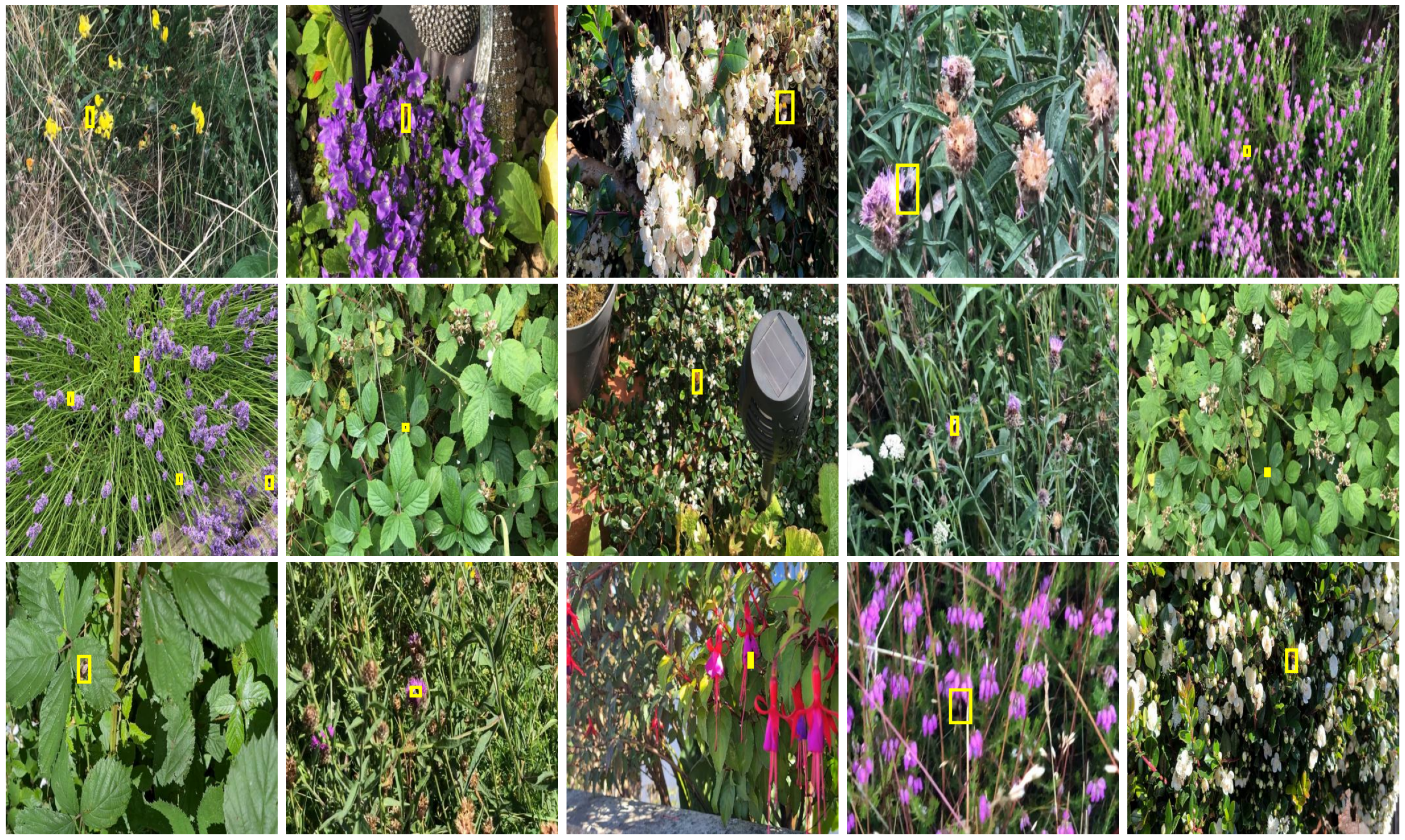}
    \caption{Examples of annotated (yellow bounding boxes) and preprocessed images within the dataset, selected at random.}
    \label{fig:examples-annots}
\end{figure}
Figure \ref{fig:examples-annots} shows examples of annotated images selected from the dataset at random. The dataset contains examples of a range of bee types, such as Bombus and Apis mellifera; recognition is based on the holistic bee, and future work could explore the recognition of individual species. The dataset generated for this study is available for public download\footnote{Dataset avaialable from:\\ \url{https://www.kaggle.com/datasets/birdy654/bee-detection-in-the-wild}}.

\subsection{Machine Learning and Data Augmentation}
Following data collection and preprocessing, several state-of-the-art object detection algorithms are then benchmarked on the \textit{Bee Detection in the Wild} dataset. This includes YOLOv5, YOLOv5m (variants of frozen and unfrozen weights), YOLOv5s, and YOLOv8m\cite{ultralytics2021yolov5}\footnote{Further details on YOLO can be found at:\\ \url{https://github.com/ultralytics/ultralytics}}.
Data augmentation for the approaches makes use of the default recommended parameters by the original authors. 
Due to differences in recommended augmentation strategies, it is important to consider data pre-processing time. Before learning on the full dataset, an initial preliminary exploration is performed into the feasibility of data augmentation; 2000 images are selected at random from the dataset and used to train YOLOv5 both with and without augmentation strategies for 30 epochs to discern whether the strategy could be useful for this problem. A preliminary test is chosen due to the constraints of computational resources, but a more in-depth exploration with the full set could be performed in the future.

Following the selection of the data preprocessing strategy, each model is trained for 100 epochs. The aforementioned validation metrics of precision, recall, mAP@.5, mAP@.5:.95, preprocessing time training time, and inference speed are measured and compared. Considerations are given to both model ability in terms of object detection, and also the computational resources required (i.e. comparison of inference time) given that real-world deployment of this model could require recognition frequency to be performed in real-time.

\subsection{Stakeholder-facing Bee Detection and Time Stamping}
\begin{figure}[]
    \centering
    \includegraphics[width=0.8\textwidth]{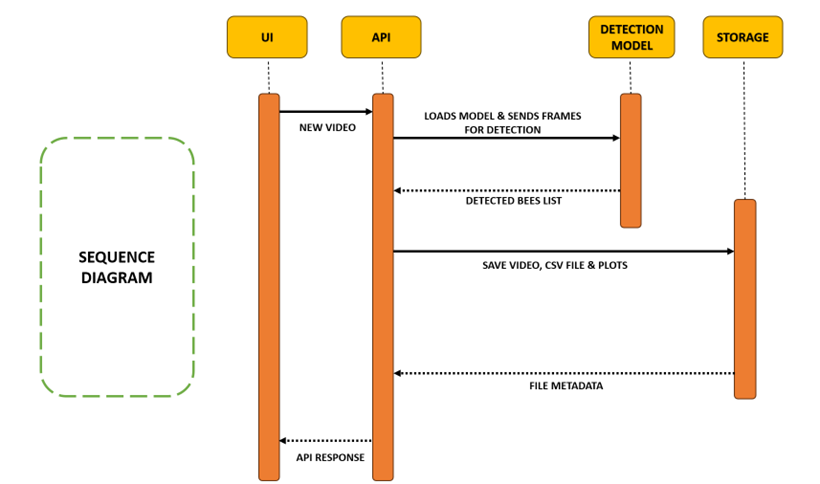}
    \caption{Graphical overview of the workflow for bee detection and timestamping.}
    \label{fig:current-system}
\end{figure}

An overview of the system can be seen in Figure \ref{fig:current-system} which describes a prototype system for bee detection and subsequent timestamping to automate the YOLO inference process. The interface is web-based and accepts video uploads (such as those collected from an outdoor camera). The system then extracts keyframes from the video at $FPS/2$, infers via the selected model, and finally distils the inference results into a report. The report contains a list of timestamps (time, bees detected) as well as visualisations in the form of charts. The system is implemented via Flask (a tool for developing Python-based web applications) and enables accessibility for key stakeholders and beneficiaries, such as bee farmers, who therefore do not require the technical ability to load the model and infer data using Python. The interaction system is open source and is available from GitHub\footnote{The source code for the system produced by this study is available from:\\ \url{https://github.com/AjayJohnAlex/Bee_Detection}}.

\section{Results and Discussion}
\label{sec:results}
This section presents the results for training and validation, as well as the testing of the object detection models on unseen data. Following that, an example of the user interface for accessibility from the stakeholder's perspective is presented, along with pertinent discussions of all the results arising from the experiments carried out in this work. 

\subsection{Preliminary Exploration}

\begin{figure}[t]
  \centering
  \begin{subfigure}[b]{0.49\textwidth}
    \includegraphics[width=\textwidth]{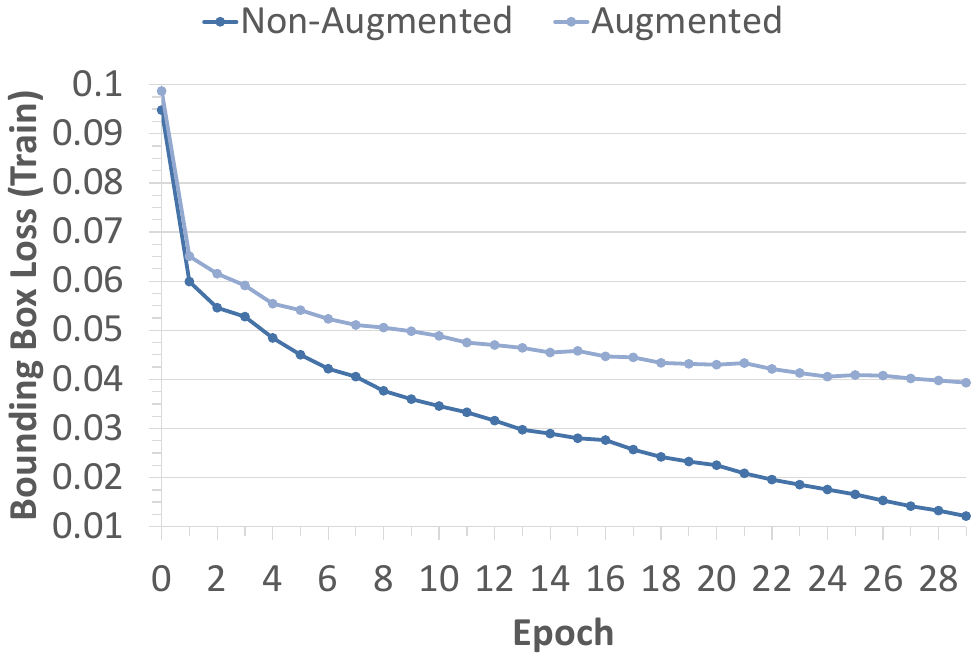}
    \caption{Bounding box loss (train data)}
    \label{fig:Train Bounding box loss}
  \end{subfigure}
  \hfill 
  \begin{subfigure}[b]{0.49\textwidth}
    \includegraphics[width=\textwidth]{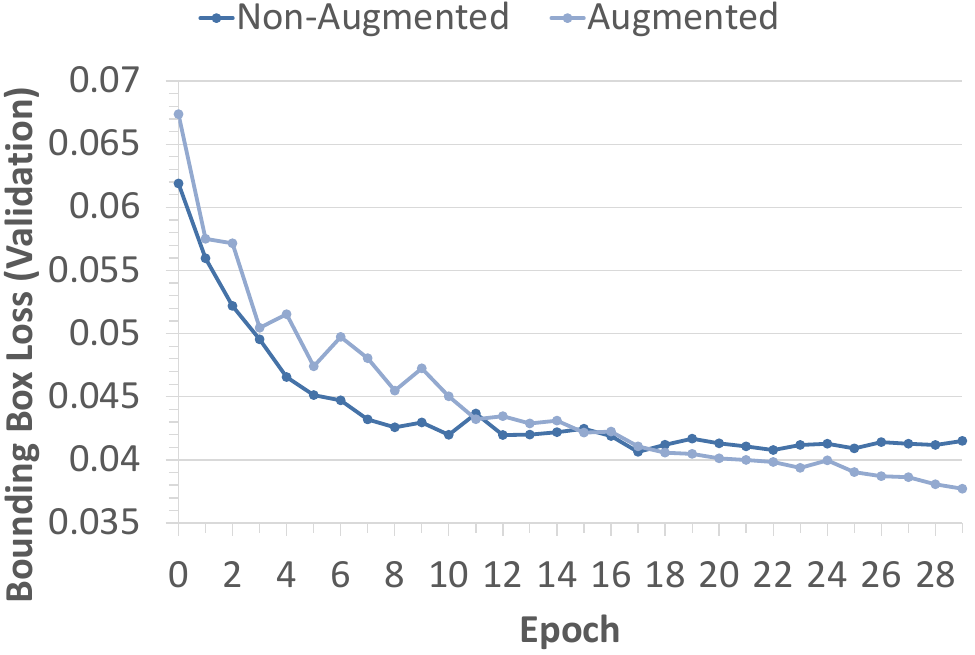}
    \caption{Bounding box loss (validation data)}
    \label{fig:Val Bounding box loss}
  \end{subfigure}
  \begin{subfigure}[b]{0.49\textwidth}
    \includegraphics[width=\textwidth]{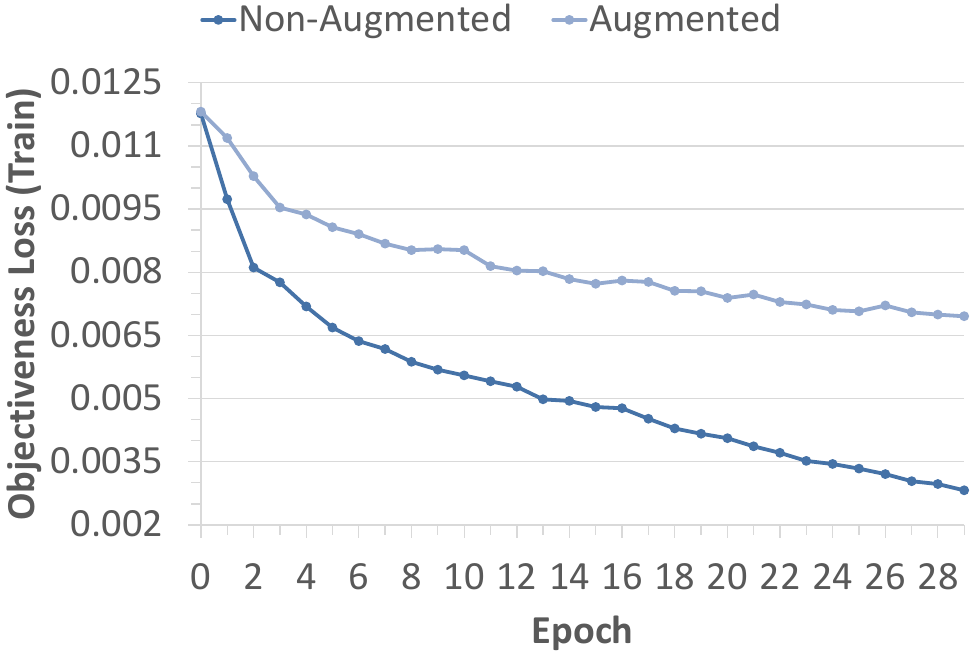}
    \caption{Objectiveness loss (train data)}
    \label{fig:Train Objectiveness loss}
  \end{subfigure}
  \hfill 
  \begin{subfigure}[b]{0.49\textwidth}
    \includegraphics[width=\textwidth]{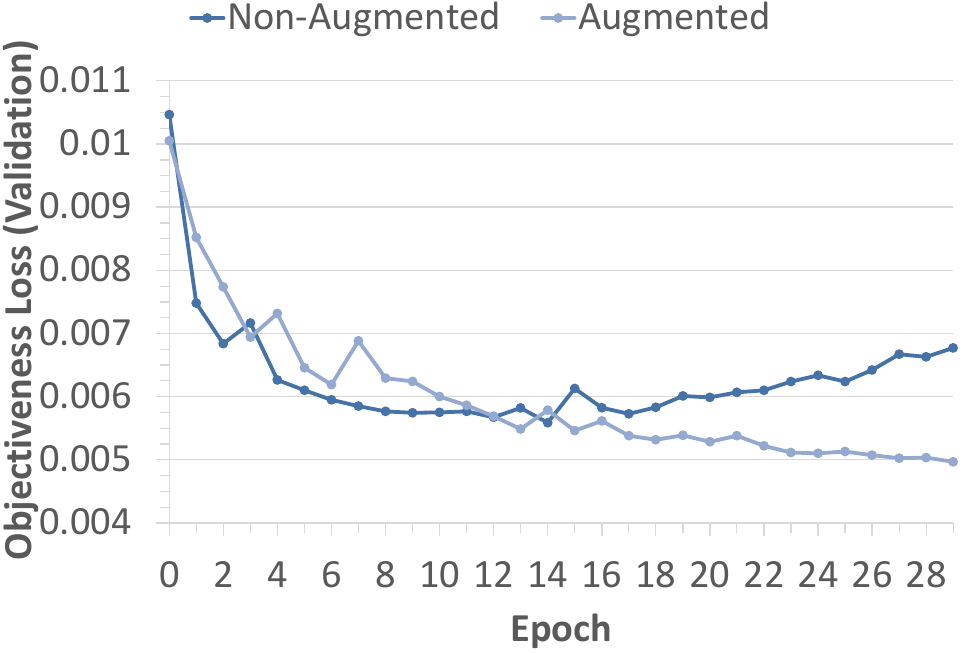}
    \caption{Objectiveness loss (validation data)}
    \label{fig:Val Objectiveness loss}
  \end{subfigure}
  \caption{Metric comparison of bounding box loss (Figures \ref{fig:Train Bounding box loss} and \ref{fig:Val Bounding box loss}) and objectiveness loss (Figures \ref{fig:Train Objectiveness loss} and \ref{fig:Val Objectiveness loss}) for a non-augmented and augmented subset of training and validation data measured over 30 epochs. \textit{Note: \mbox{y-axis} scales are not comparable.}}
  \label{fig:dataaug-1}
\end{figure}

\begin{figure}[t]
  \centering
  \begin{subfigure}[b]{0.49\textwidth}
    \includegraphics[width=\textwidth]{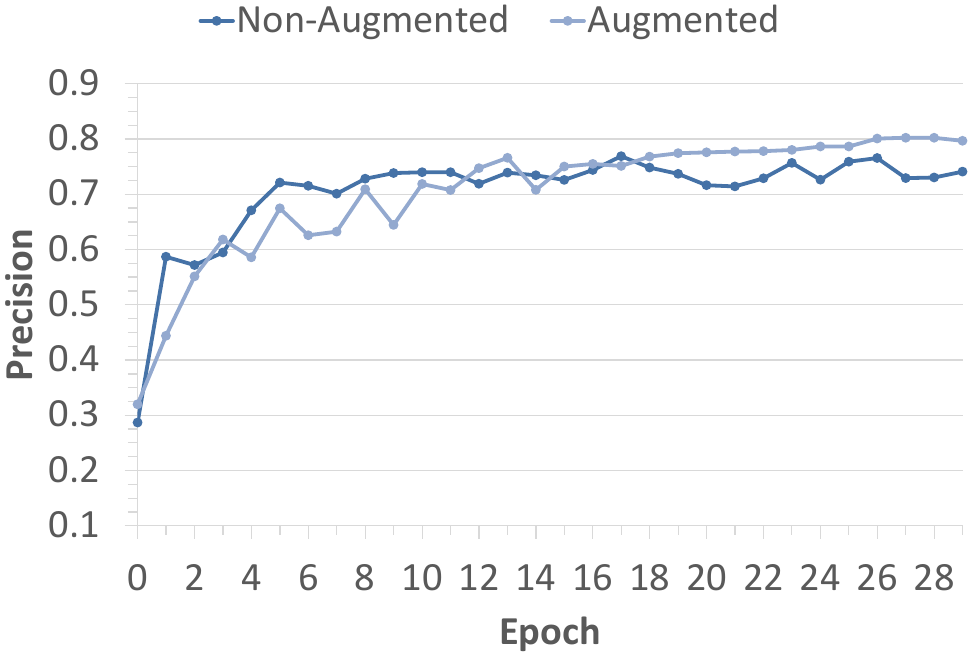}
    \caption{Precision}
    \label{fig:Precision}
  \end{subfigure}
  \hfill 
  \begin{subfigure}[b]{0.49\textwidth}
    \includegraphics[width=\textwidth]{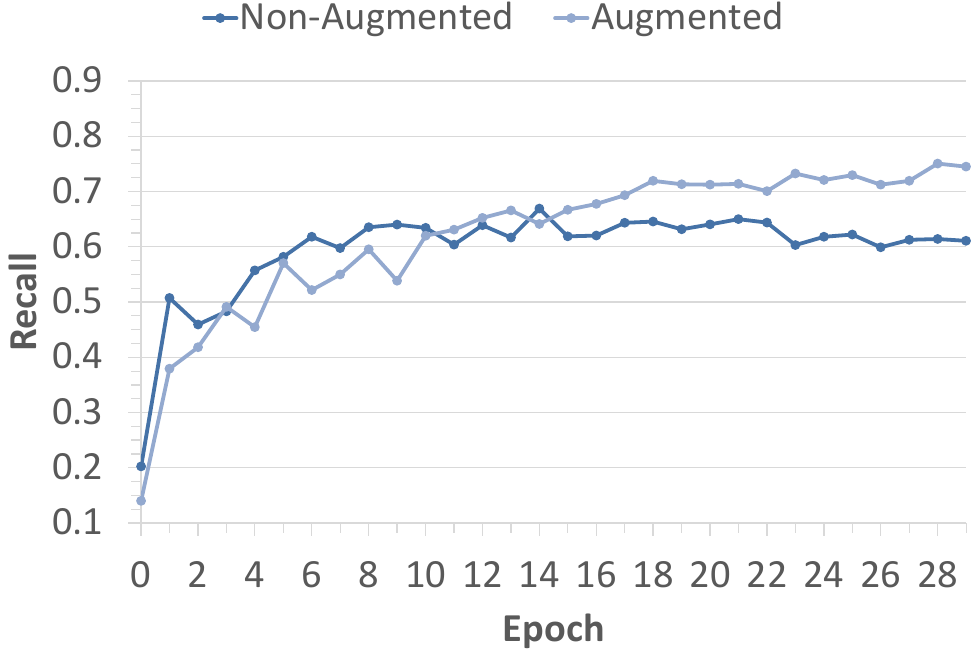}
    \caption{Recall}
    \label{fig:Recall}
  \end{subfigure}
  \begin{subfigure}[b]{0.49\textwidth}
    \includegraphics[width=\textwidth]{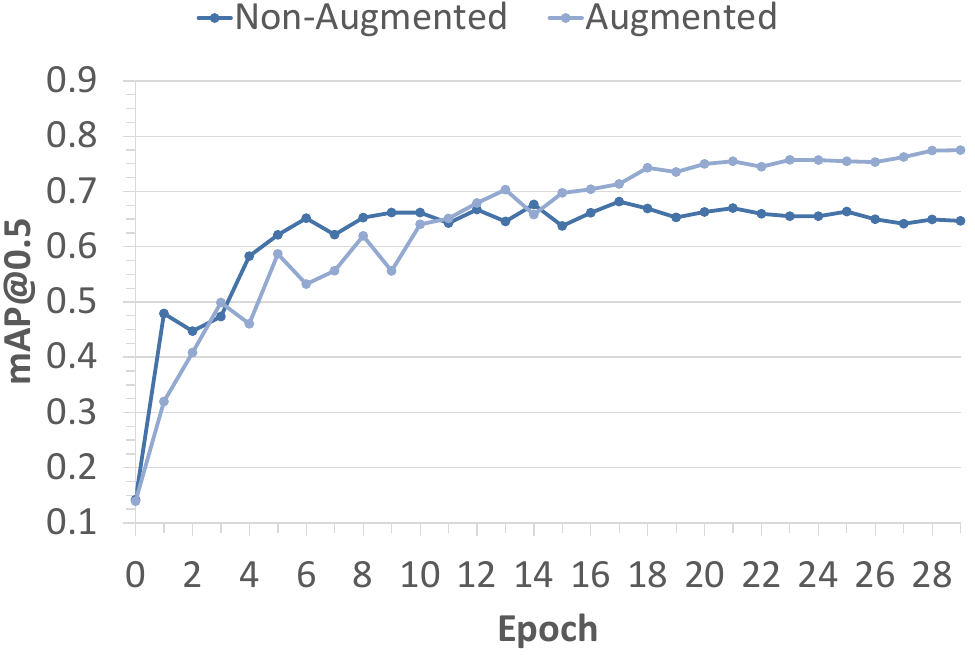}
    \caption{mAP@0.5}
    \label{fig:Accuracy_at_50}
  \end{subfigure}
  \hfill 
  \begin{subfigure}[b]{0.49\textwidth}
    \includegraphics[width=\textwidth]{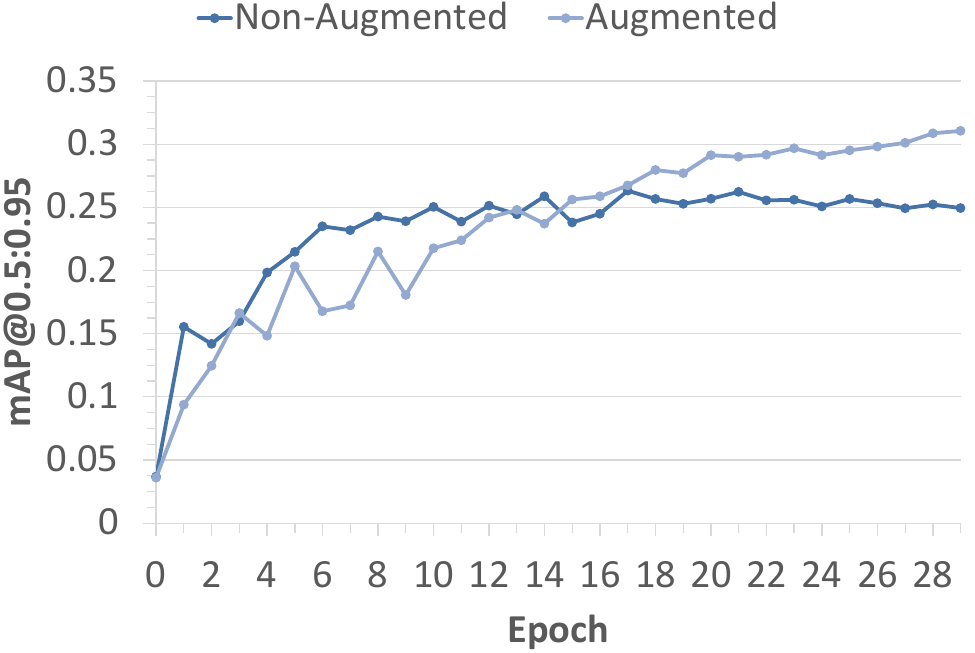}
    \caption{mAP@0.5:0.95}
    \label{fig:Accuracy_at_50_95}
  \end{subfigure}
  \caption{Metric comparison of \subref{fig:Precision}) precision, \subref{fig:Recall}) recall, \subref{fig:Accuracy_at_50}) mAP@0.5, and \subref{fig:Accuracy_at_50_95}) mAP@0.5:0.95, for a non-augmented and augmented subset of training and validation data measured over 30 epochs. \textit{Note: \mbox{y-axis} scales for Figures \ref{fig:Precision} to \ref{fig:Accuracy_at_50} are not comparable to Figure \ref{fig:Accuracy_at_50_95}.}}
  \label{fig:dataaug2}
\end{figure}

An example of data augmentation and its effect on training using a subset of data can be seen in Figures \ref{fig:dataaug-1} and \ref{fig:dataaug2}. Validation losses for both the bounding box (Figure~\ref{fig:Val Bounding box loss}) and objectiveness (Figure~\ref{fig:Val Objectiveness loss}) are lower when augmentation strategies are implemented. After 30 epochs using non-augmented and augmented validation data, the bounding-box losses were 0.0415 and 0.0377 respectively (to 3 S.F.). Similarly, for objectiveness loss, non-augmented and augmented validation data losses were 0.00677 and 0.00497 respectively (to 3 S.F.). These results demonstrate the effectiveness of data augmentation during validation.

Interestingly, this effect cannot be seen during training, where the augmented losses are higher for the same metrics (Figures~\ref{fig:Train Bounding box loss} and \ref{fig:Train Objectiveness loss} respectively). After 30 epochs, training on nonaugmented and augmented data lead to a bounding-box loss of 0.0122 and 0.0393 respectively, with 0.00282 and 0.00695 reported for objectiveness loss (to 3 S.F.).\\

\subsection{Training Results}

\begin{table}[]
\centering
\caption{YOLOv5 model training summary}
\label{tab:yolov5-results}
\footnotesize
\begin{tabular}{ccccc}
\toprule
\textbf{Metric} & \textbf{YOLOv5s} & \textbf{YOLOv5m} & \textbf{YOLOv5m-frozen} & \textbf{Base YOLOv5} \\
\midrule
\textbf{\textit{Precision}} & 81.4 & 81.0 & 75.5 & 80.0 \\
\textbf{\textit{Recall}} & 78.6 & 80.8 & 64.8 & 72.1 \\
\textbf{\textit{mAP@.5}} & 80.9 & 81.5 & 69.0 & 76 \\
\textbf{\textit{mAP@.5:.95}} & 34.2 & 35.6 & 27.5 & 30.6 \\
\textbf{\textit{Inference Speed (ms)}} & 6.3 & 7.6 & 8.3 & 6.9 \\
\textbf{\textit{GPU Training Time (min)}} & 124 & 163  & 128  & 119  \\
\bottomrule
\end{tabular}
\end{table}

A summary of the training results of the YOLOv5 series of models can be found in Table \ref{tab:yolov5-results}. It was observed that YOLOv5m has a lower precision than YOLOv5s (81.0 and 81.4, respectively); however, it experiences higher mAP scores for both related metrics (81.5 and 35.6 compared to 80.9 and 34.2, respectively). YOLOv5m with frozen weights has significantly lower recall and mAP values at 64.8, 69.0, and 27.5, respectively. This suggests that the dataset diverges from COCO-related detection activities. Although the Base YOLOv5 model was the quickest to train, the YOLOv5s has a much lower training inference time of 6.3ms, with a higher speed suggesting that it is more suitable for real-time application. 

\begin{table}[]
\centering
\caption{Indirect training comparison between YOLOv5 variants with YOLOv8}
\label{tab:indirect-comp}
\footnotesize
\begin{tabular}{ccccc}
\toprule
\textbf{Model Summary} & \textbf{YOLOv5s} & \textbf{YOLOv5m} & \textbf{YOLOv8} \\
\midrule
\textbf{\textit{Precision}} & 81.4 & 81.0 & 81.8 \\
\textbf{\textit{Recall}} & 78.6 & 80.8 & 80.4 \\
\textbf{\textit{mAP@.5}} & 80.9 & 81.5 & 83.3 \\
\textbf{\textit{mAP@.5:.95}} & 34.2 & 35.6 & 37.8 \\
\textbf{\textit{Inference Speed (ms)}} & 6.3 & 7.6 & 4.2 \\
\textbf{\textit{GPU Training Time (min)}} & 124 & 163  & 214 \\
\bottomrule
\end{tabular}
\end{table}

The results in Table \ref{tab:indirect-comp} show a comparison between the YOLO models v5 and v8. Since both follow their default, recommended, and different augmentation strategies, the comparison is indirect. Although YOLOv8 has a lower recall score than YOLOv5m, all other metrics outperform it; this includes precision, both mAP scores, and a lower inference speed. It is also worth noting that YOLOv8 takes a considerably longer time to train, suggesting a larger initial investment in computational resources before improving real-time detection ability.

\subsection{Testing Results}

\begin{table}[]
\centering
\caption{Object recognition ability for the models on the testing data. Inference denotes the total average time taken to fully process an input frame.}
\label{tab:testing-results}
\footnotesize
\begin{tabular}{@{}llllll@{}}
\toprule
\textbf{Model}                   & \textbf{Precision} & \textbf{Recall} & \textbf{mAP@.5} & \textbf{mAP@.5:.95} & \textbf{Inference (ms)} \\ \midrule
\textit{\textbf{Base YOLOv5}}    & 81.5               & 76.6            & 81.4            & 38.3                & 4.8                      \\
\textit{\textbf{YOLOv5s}}        & 82.6               & 79.7            & 84.6            & 41                  & 5.1                      \\
\textit{\textbf{YOLOv5m}}        & 83.1               & 81.4            & 85.6            & 42.2                & 8.1                      \\
\textit{\textbf{YOLOv5m-frozen}} & 77.2               & 67.5            & 75.9            & 34.7                & 8.5                      \\
\textit{\textbf{YOLOv8m}}        & 81.9               & 80.3            & 83.3            & 37.8                & 12.5                     \\ \bottomrule
\end{tabular}
\end{table}

\begin{figure}[]
  \centering
  \begin{subfigure}[b]{0.49\textwidth}
    \includegraphics[width=\textwidth]{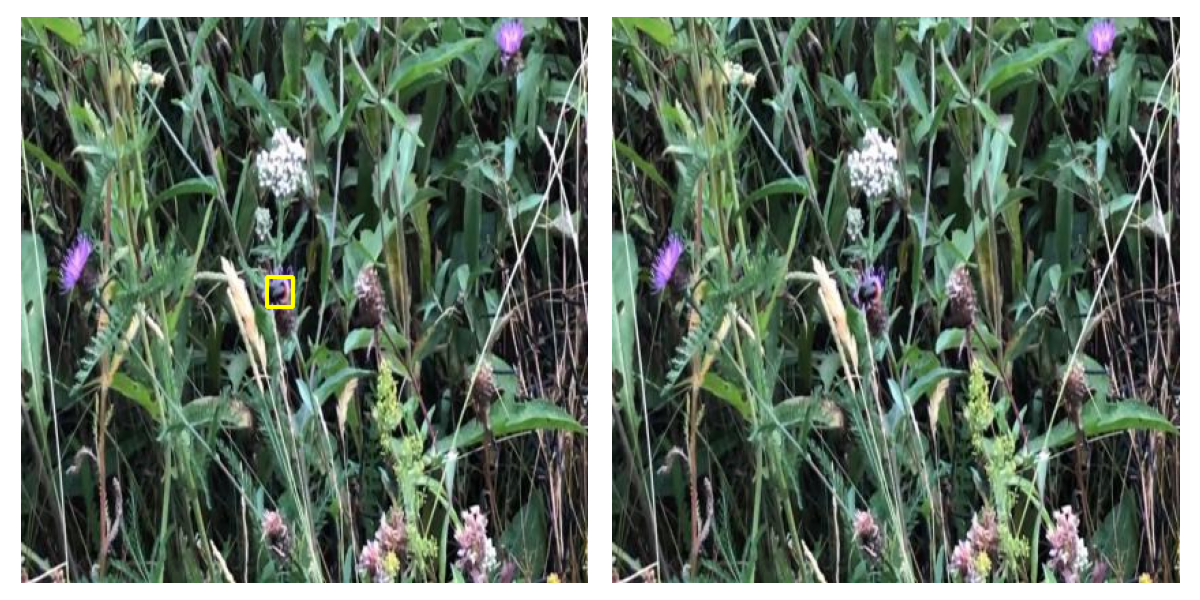}
    \caption{}
    \label{subfig:mistake-1}
  \end{subfigure}
  \hfill 
  \begin{subfigure}[b]{0.49\textwidth}
    \includegraphics[width=\textwidth]{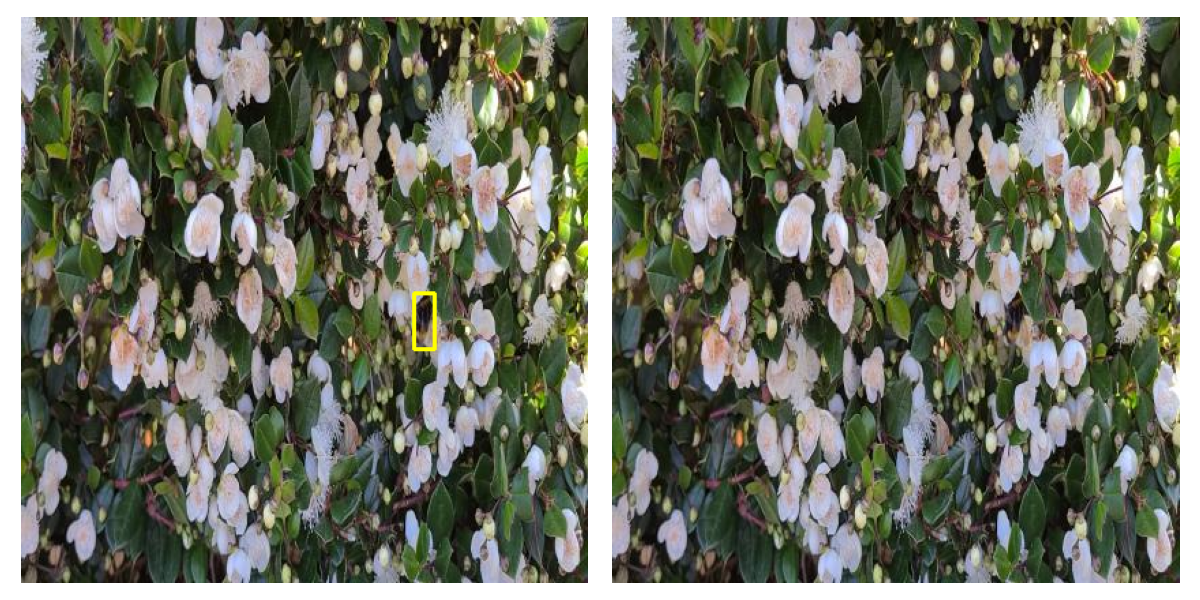}
    \caption{}
    \label{subfig:mistake-2}
  \end{subfigure}
  \begin{subfigure}[b]{0.49\textwidth}
    \includegraphics[width=\textwidth]{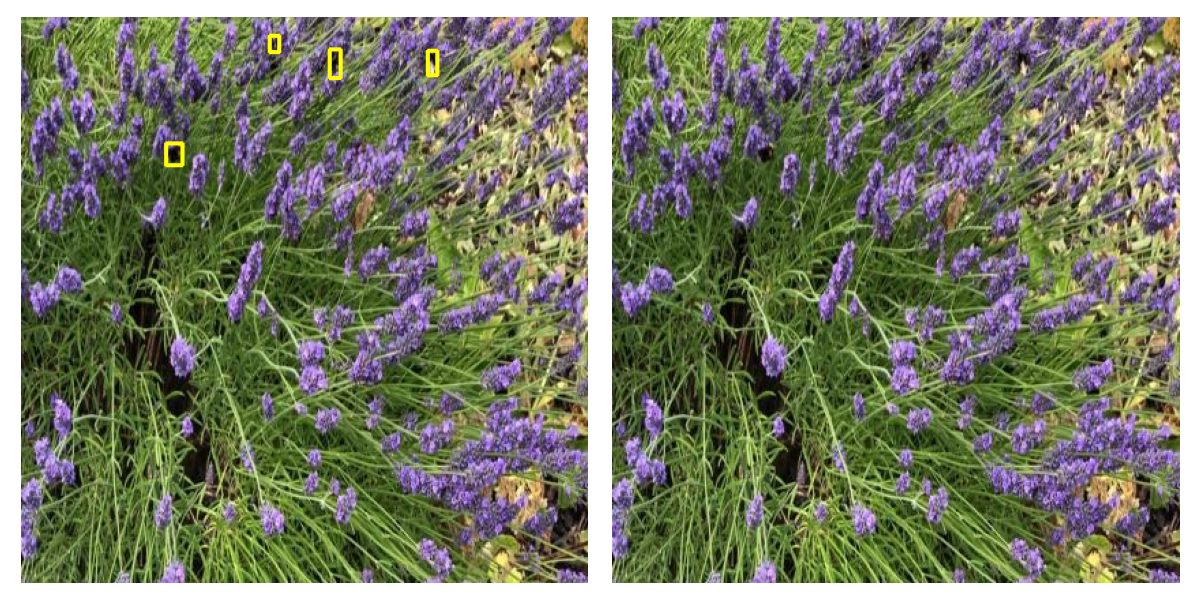}
    \caption{}
    \label{subfig:mistake-3}
  \end{subfigure}
  \hfill 
  \begin{subfigure}[b]{0.49\textwidth}
    \includegraphics[width=\textwidth]{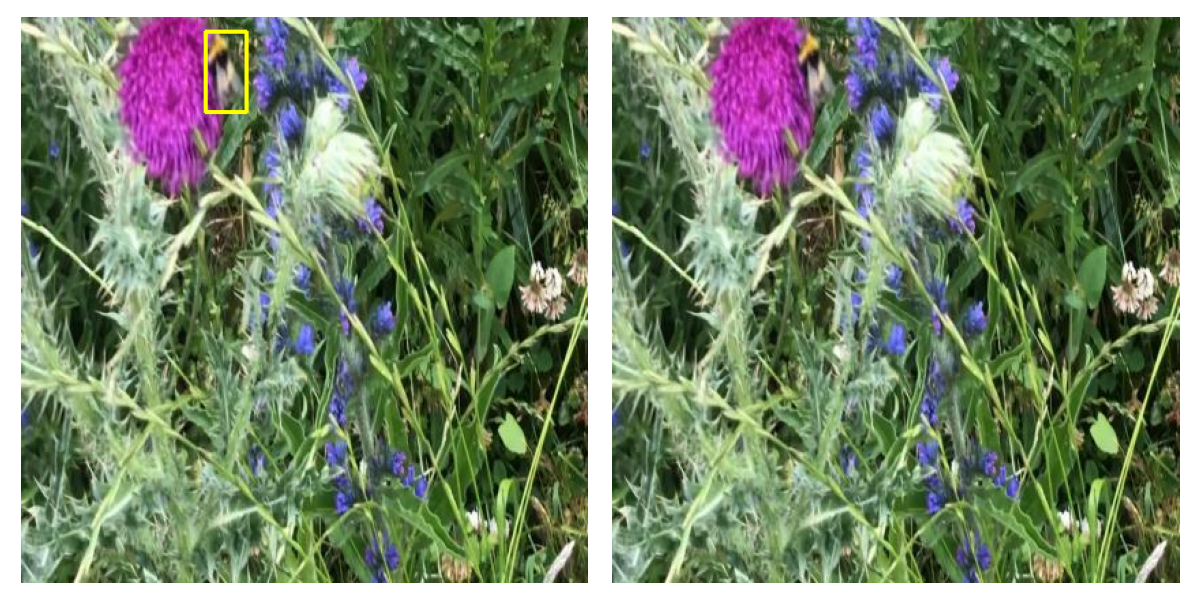}
    \caption{}
    \label{subfig:mistake-4}
  \end{subfigure}
  \caption{Examples of ground-truth bees missed by the YOLOv5m model. Left images show ground truth data with yellow bounding boxes, right image shows a lack of model predictions.}
  \label{fig:mistakes}
\end{figure}

The results in Table \ref{tab:testing-results} show the test results for each of the models. Given these data, it was observed that YOLOv5m has the highest precision at 83.1\%, suggesting that it was the best model to identify relevant objects. This approach also scored the highest recall, so it missed the fewest bees when making recognition and classification predictions. At both mAP values, YOLOv5m also scored the highest. Figure \ref{fig:mistakes} shows examples of mistakes made by the YOLOv5m model. The results show that, even with a precision of 83.1\%, the model can miss both relatively small bees (Figures \ref{subfig:mistake-1} and \ref{subfig:mistake-3}) and relatively large bees (Figures \ref{subfig:mistake-2} and \ref{subfig:mistake-4}). To this end, examples of misclassified data suggest that additional data could be collected in the future to further improve the robustness of the approach.

\begin{table}[]
\centering
\caption{Breakdown of average processing time for the models on the testing data}
\label{tab:speed-comparison}
\footnotesize
\begin{tabular}{@{}lllll@{}}
\toprule
\multirow{2}{*}{\textbf{Model}}  & \multicolumn{4}{l}{\textbf{Average processing time (ms)}}                                                     \\ \cmidrule(l){2-5} 
                                 & \textit{\textbf{Pre-process}} & \textit{\textbf{Inference}} & \textit{\textbf{NMS}} & \textit{\textbf{Total}} \\ \cmidrule(r){1-1}
\textit{\textbf{Base YOLOv5}}    & 0.1                           & 3.4                         & 1.3                   & 4.8                     \\
\textit{\textbf{YOLOv5s}}        & 0.1                           & 3.4                         & 1.6                   & 5.1                     \\
\textit{\textbf{YOLOv5m}}        & 0.1                           & 6.7                         & 1.3                   & 8.1                     \\
\textit{\textbf{YOLOv5m-frozen}} & 0.1                           & 6.9                         & 1.5                   & 8.5                     \\
\textit{\textbf{YOLOv8m}}        & 0.8                           & 10.3                        & 1.4                   & 12.5                    \\ \bottomrule
\end{tabular}
\end{table}

Real-time object detection, of paramount importance when transferring research from the laboratory to real-world actionable insights, is the ability to perform inference at suitable speeds as outlined by the problem it is designed to solve; for example, a model that may contribute to a critical decision (such as an autonomous vehicle) should have a low latency time to avoid causing an emergency. In apiculture, a low inference time for bee detection is important for real-time monitoring and responses to changes in activity or health, as well as lower computational complexity making the algorithms more accessible to those who may not have access to high-end computing resources. Table \ref{tab:speed-comparison} shows a more extensive set of results for time-based metrics. In these experiments, the testing dataset of 996 images was used as input for prediction. The difference in default and recommended augmentation strategies can be seen in the pre-processing time, where YOLOv8 takes 0.7 seconds compared to 0.3. As also observed, YOLOv8m has the highest inference time, with the lowest being the Yolov5 and YOLOv5s models. Non-Maximum Suppression (NMS) was relatively similar across all models, with the lowest being 1.3ms (V5 and V5m) and the highest was 1.6ms (V5s). Overall, the quickest model was Base YOLOv5 at 4.8ms, closely followed by V5s at 5.1ms. The highest performing model, YOLOv5m, took longer at 8.1ms per frame.

\subsection{Stakeholder-facing Interface}
As suggested during the literature review, the use of video streams is showing promise for autonomous monitoring of pollinator behaviour. They are non-intrusive and can be used to capture critical data such as hive health and population decline. Given that Python code and machine learning models are often presented in technical formats, they are therefore inaccessible to a wider audience. Following the training and validation of the models, this article proposes an encapsulation of the work in a format that is accessible to the stakeholder. 

\begin{figure}[]
    \centering
    \includegraphics[scale=0.4]{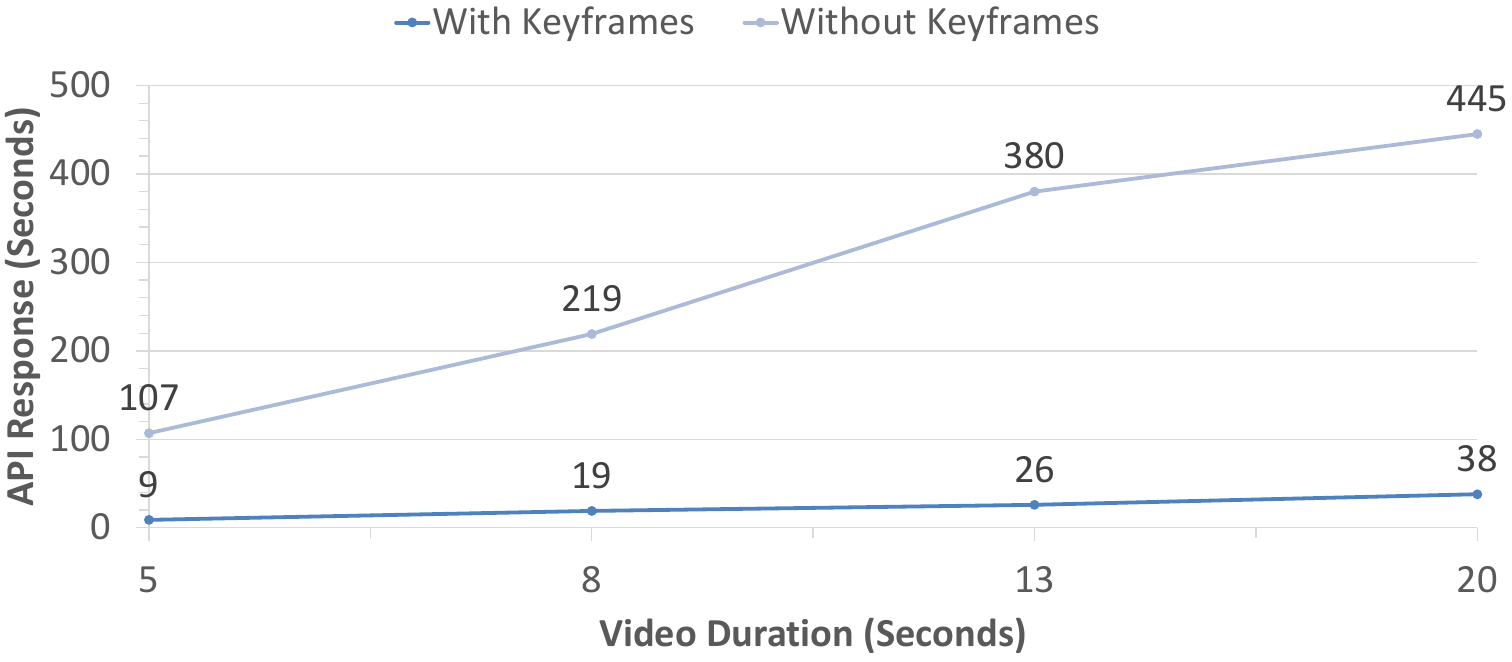}
    \caption{Impact of keyframe selection on API Response}
    \label{fig:keyframes}
\end{figure}

Figure \ref{fig:keyframes} shows the impact of keyframe selection on API response time. As expected, when frames per second are reduced by half, that is, $FPS/2$ keyframes from video streams, the processing time is reduced. Videos of 5, 8, 13, and 20 seconds could be processed in 9, 19, 26, and 38 seconds, respectively. In the future, keyframe selection strategies could be benchmarked according to the relevant literature, which is discussed in Section \ref{sec:conclusion}. 

\begin{figure}
    \centering
    \includegraphics[width=\textwidth]{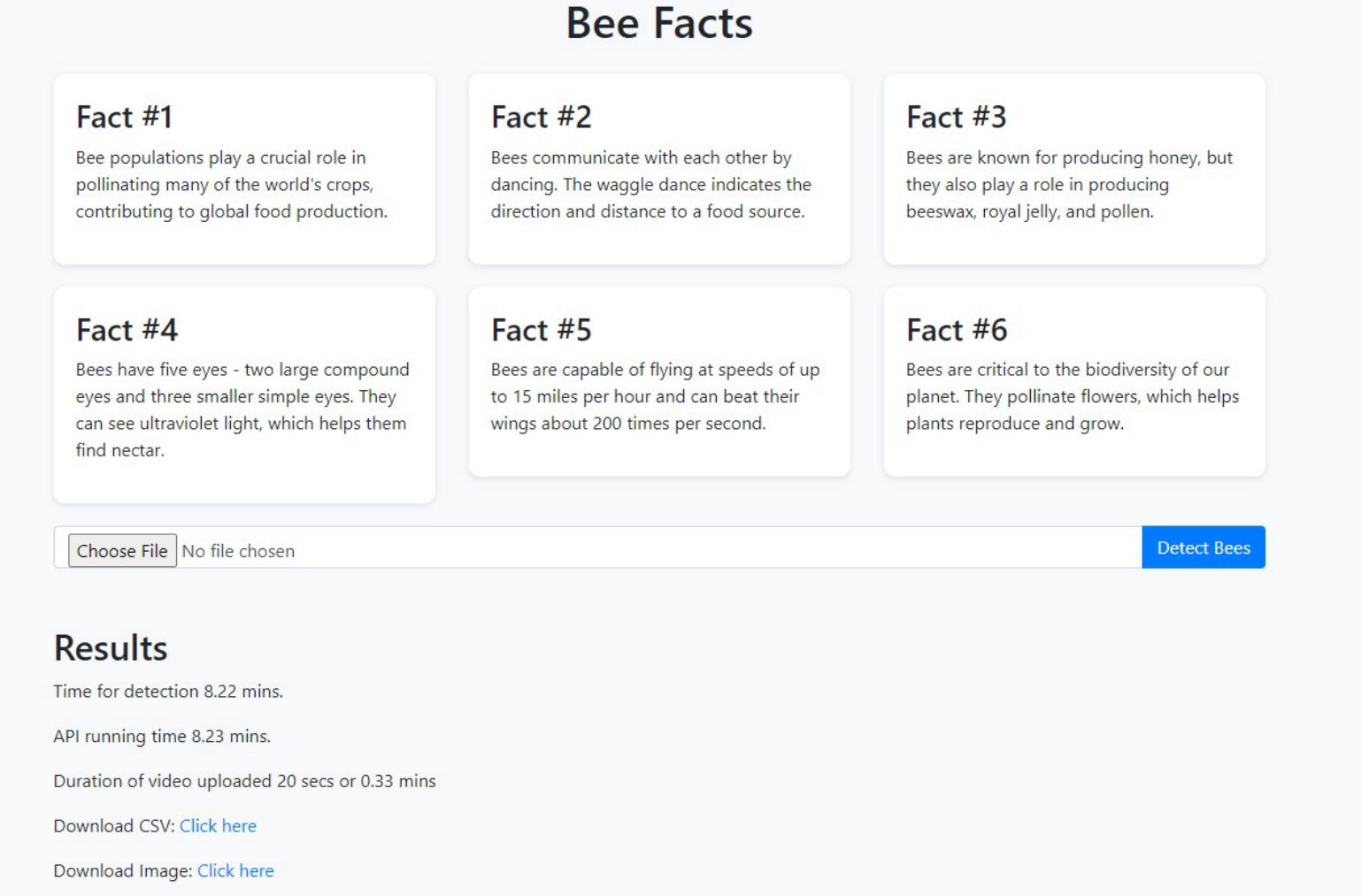}
    \caption{A screenshot of the interface for the non-technical interaction with the inference model. For demonstration purposes, random bee facts are printed to the screen.}
    \label{fig:UI}
\end{figure}

An example of the user interface can be found in Figure \ref{fig:UI}. The system encapsulates all model inference and replaces the need to either write Python code or use the command-line interface by enabling the upload of a video file. This video is then automatically processed by the object detection model and a report is generated along with images of the detected bees for viewing and further analysis. For demonstration purposes, random bee facts are printed onto the screen, but this can be replaced or removed entirely in the future. 

\begin{table}[]
\centering
\caption{An excerpt of an example report showing the event of a bee entering the camera view and being detected by the object detection model. }
\label{tab:example-report}
\begin{tabular}{@{}llllllll@{}}
\toprule
\textbf{D} & \textbf{H} & \textbf{M} & \textbf{S} & \textbf{\begin{tabular}[c]{@{}l@{}}Video\\ Time\end{tabular}} & \textbf{Time (formatted)}      & \textbf{DD\_HH\_MM\_SS} & \textbf{Detected} \\ \midrule
0          & 0          & 15         & 59         & 959                                                           & 0 days 0 hours 15 mins 59 secs & 00:00:15:59             & 0                 \\
0          & 0          & 16         & 0          & 960                                                           & 0 days 0 hours 16 mins 0 secs  & 00:00:16:00             & 0                 \\
0          & 0          & 16         & 1          & 961                                                           & 0 days 0 hours 16 mins 1 secs  & 00:00:16:01             & 0                 \\
0          & 0          & 16         & 2          & 962                                                           & 0 days 0 hours 16 mins 2 secs  & 00:00:16:02             & 0                 \\
0          & 0          & 16         & 3          & 963                                                           & 0 days 0 hours 16 mins 3 secs  & 00:00:16:03             & 0                 \\
0          & 0          & 16         & 4          & 964                                                           & 0 days 0 hours 16 mins 4 secs  & 00:00:16:04             & 0                 \\
0          & 0          & 16         & 5          & 965                                                           & 0 days 0 hours 16 mins 5 secs  & 00:00:16:05             & 0                 \\
0          & 0          & 16         & 6          & 966                                                           & 0 days 0 hours 16 mins 6 secs  & 00:00:16:06             & 1                 \\
0          & 0          & 16         & 7          & 967                                                           & 0 days 0 hours 16 mins 7 secs  & 00:00:16:07             & 1                 \\
0          & 0          & 16         & 8          & 968                                                           & 0 days 0 hours 16 mins 8 secs  & 00:00:16:08             & 1                 \\
0          & 0          & 16         & 9          & 969                                                           & 0 days 0 hours 16 mins 9 secs  & 00:00:16:09             & 1                 \\
0          & 0          & 16         & 10         & 970                                                           & 0 days 0 hours 16 mins 10 secs & 00:00:16:10             & 1                 \\
0          & 0          & 16         & 11         & 971                                                           & 0 days 0 hours 16 mins 11 secs & 00:00:16:11             & 1                 \\ \bottomrule
\end{tabular}%
\end{table}

Following the submission and processing of a video stream, an example of the aforementioned report can be found in Table \ref{tab:example-report}. These rows are exemplars extracted from the CSV file generated, and show how at 16:06, the model predicted that a bee had entered the frame and thus received a bounding box. It is important to note that this report was generated solely by following three steps: 1) clicking the ``Choose File" button, 2) choosing a video file, and 3) clicking the ``Detect bees" button. No interaction with code or the command line interface is required, with the aim of democratising the technology arising from the scientific contributions of this work.

\section{Conclusion and Future Work}
\label{sec:conclusion}
This study has contributed a significantly large dataset for the research community and explored the application of object detection algorithms for bee detection and tracking. The use of these data and/or these algorithms are part of an endeavour to provide a technological intervention in conservation management and analysis. It is particularly pertinent to explore these interventions for pollinators given that food security and the overall survival of the human species is dependent on their services. The democratisation of research is also supported by the approaches proposed by this work, given the open-source nature of all the data, models and software produced. The findings showed that two particular models benchmarked in this work are promising, the YOLOv5m model which achieved 85.6\& mAP@0.5 with an inference time of 8.1ms per frame, and the more efficient YOLOv5s at 5.1ms inference at a slightly lower 84.6\% mAP. 

Beyond data collection and machine learning implementation, this work also argued in favour of stakeholder-friendly and accessible application, and contributed to this in the form of a web application for automated analysis using the selected model. This meant that operators did not need to use any computer code to analyse video streams, such as those collected from an apiary. 

A limitation of this work is the preliminary nature of the selection of the data augmentation strategy, which was restricted to a data subset due to the availability of computational resources. In the future, multiple augmentation strategies could be explored using the full dataset to discern a more general view of which is most effective to use. This, along with combinatorial optimisation of the augmentation parameters, could lead to a better overall model. In terms of real-time processing, a relatively simple keyframe selection strategy of $FPS/2$ was implemented; however, this process could be improved by making use of a more in-depth keyframe selection strategy according to the literature\cite{rashmi2018effective,huang2019novel,savran2023novel}.

\begin{figure}[]
    \centering
    \includegraphics[scale=0.8]{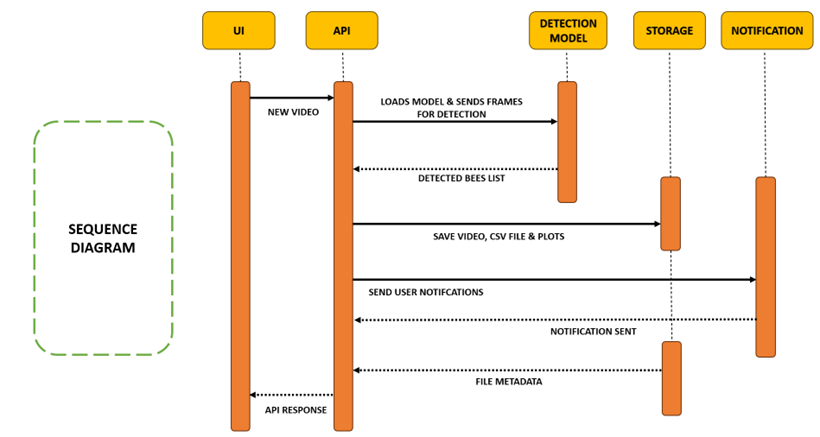}
    \caption{Proposed future pipeline for improved use in industry.}
    \label{fig:future-workflow}
\end{figure}

In the future, the application of these models in real-world situations, such as an apiary, could provide valuable insight into behavioural analysis. Furthermore, multimodality could improve detection techniques, such as the fusion of audio collected alongside images. Future work could also explore multiple classes, such as the detection of individual species, or behaviours, such as the automatic recognition of pollination events. If used in conjunction with a flower recognition algorithm, this algorithm could then enhance the reporting system towards more detailed information about events that have taken place. Toward the next iteration of this work, Figure \ref{fig:future-workflow} shows an example of how the pipeline could be improved. This includes sending notifications to stakeholders and beneficiaries (such as an alert to a population decline) along with the transition away from local to cloud storage. Additionally, the user interface printed random bee facts for demonstration purposes; improvements could be made to the interface in future, such as replacing this block with a live feed of the detection process, that is, by showing images of some of the bees detected and live graphs that update as the inference process is performed. 

Toward integration reflective of Agriculture 4.0, future implementations of this framework could consider wireless communication, such as inference on the edge and wireless communication (5G, 6G, LoraWan etc.). Additionally, enhancements through Geographic Information System (GIS) and Global Positioning System (GPS) data could enable the provision of further context through geographic and location metadata. Furthermore, given long-term data collection and modelling, findings could be used to infer design decisions when building digital twins of the relevant ecosystems. 

To finally conclude, this work has proposed the use of a novel dataset and object detection algorithms to facilitate technological intervention in conservation management, with a focus on pollinator behaviour. Food security, reacting to climate change, the conservation of the natural world, and responsible consumption and production are critical issues that support human survival, and the use of technology to monitor their improvement is possible given the transfer of interdisciplinary knowledge from the laboratory to the real world. 

\section{Data Availability Statement}
All data collected and subsequent code written is made publicly available for future work. 

The \textit{Bee Detection in the Wild}, collected and analysed in this study, is released via the Kaggle data science platform under the MIT license. It can be downloaded from: \url{https://www.kaggle.com/datasets/birdy654/bee-detection-in-the-wild}

The code for the web interface used to encapsulate the models is available on Github. It can be downloaded from: \url{https://github.com/AjayJohnAlex/Bee_Detection}

\bibliographystyle{unsrt} 
\bibliography{references}

\end{document}